\pdfoutput=1

\documentclass{article}
\usepackage{url}
\usepackage[utf8]{inputenc} 
\usepackage[T1]{fontenc}    
\usepackage{authblk}

\usepackage[preprint]{neurips2021}

\usepackage{url}
\usepackage{verbatim}
\usepackage{hyperref}
\usepackage{times}
\usepackage{latexsym}
\usepackage{amsmath}
\usepackage{graphicx}
\usepackage{tikz}
\usepackage{tikz-dependency}
\usepackage{subcaption}
\usepackage{todonotes}
\usepackage{booktabs}
\usepackage{enumitem}
\usepackage{xcolor}
\usepackage{amsfonts}

\usepackage{comment}
\usepackage{natbib}
\usepackage{fancyhdr}
\usepackage{nicefrac}       
\usepackage{microtype}      

\title{Connecting Neural Response measurements \& Computational Models of language: a non-comprehensive guide} 

\author{%
  \textbf{Mostafa Abdou} \\
  University of Copenhagen \\
  \texttt{abdou@di.ku.dk} \\
}

\begin{document}
\maketitle
\begin{abstract}
Understanding the neural basis of language comprehension in the brain has been a long-standing goal of various scientific research programs. Recent advances in language modelling and in neuroimaging methedology promise potential improvements in both the investigation of language's neurobiology and in the building of better and more human-like language models. This survey traces a line from early research linking Event Related Potentials and complexity measures derived from simple language models to contemporary studies employing Artificial Neural Network models trained on large corpora in combination with neural response recordings from multiple modalities using naturalistic stimuli. 
\end{abstract}

\section{Introduction}
The mechanisms which underlie language comprehension in humans have been the object of study of a broad range of scientific research programs. Early work in theoretical and psycho- linguistics hypothesized certain mechanisms and structures underlying language processing, often employing behavioural data to confirm or refute them \citep{taylor1953cloze, chomsky2009syntactic, chomsky2014aspects, chomsky2014minimalist,greenberg1963some, katz1963structure, fodor1966some, lenneberg1967biological, geschwind1970organization, lakoff2008metaphors, fodor1983modularity, mcclelland1986parallel, prince2008optimality, luce1998recognizing, rayner1998eye}, inter alia. With the advancement of neuroimaging technologies, it became possible to begin to localise some of the computations responsible for language in the brain --- both in time and space --- investigating a) the timeline of such computations and b) the brain regions (or networks) which carry them out. 

Orthogonally, advances in the field of artificial intelligence have enabled the training and deployment of large artificial neural network models (ANNs). These models, which are loosely-based on upon the structure of biological brains (Haykin, 1994), have demonstrated remarkable adeptness at a wide variety of tasks \citep{bengio2009learning, krizhevsky2012imagenet, graves2013speech,schmidhuber2015deep, silver2016mastering,  goodfellow2016deep}. In the field of natural language processing (NLP), ANNs have all but replaced other types of statistical methods previously employed, showing far superior performance on an assortment of natural language understanding tasks \citep{devlin2018bert, radford2019language, brown2020language}. 

Although they fundamentally differ from the neural architecture of the human brain, the success of these models in approximating human behaviour on tasks such as object recognition and speech recognition and in various language understanding tasks led to the suggestion that they could be adopted as potential models of the representations and structures which underpin human cognition. In seminal work, researchers found that convolutional neural networks \citep{lecun1995convolutional} trained on large image classification datasets could predict image-evoked neural activations in the ventral visual stream with a higher accuracy than all previous models --- even those directly optimised to fit neural activations \citep{yamins2014performance, yamins2016using}. Similar research followed for a variety of perceptual domains, showing that ANNs exhibit similarities to both human behaviour and neural responses and that those similarities arise simply as a consequence of learning to perform a task such as image classification \citep{kriegeskorte2015deep, eickenberg2017seeing, kell2018task}. Analogously, for language, ANN models trained to predict future or masked words from context have recently been found to show considerable representational alignment to neural response the human brain \citep{caucheteux2020language, schrimpf2020integrative, goldstein2021thinking}.

\paragraph{Scope}
This paper makes a survey of research which links computational models of and neural responses to language. Particular attention is given to more recent work which makes use of ANNs, with the goal of tracing a line between this trend and previous work leveraging simpler computational models such as syntactic parsing models or context-free grammars. Studies which combine those, or any type of computational model, with neuroimaging data are considered to be within the scope of the survey, while those e.g. primarily focusing on the stimuli themselves are not. For other relevant surveys with different foci, readers are referred to \cite{murphy2018decoding} and \cite{hale2021neuro}.

\paragraph{Overview}
The rest of this paper is structured as follows: Section \ref{sec:prel} outlines a set of preliminaries, establishing terminology that will be used throughout the paper; Section \ref{sec:old_studies} offers historical context; Section \ref{sec:meaning_nouns} describes early work which focused on representations of and neural responses to words, treated as isolated units; Section \ref{sec:syntax} surveys research that employs computational models in investigations of syntactic structure in the brain; Section \ref{sec:multiple} moves forward to studies which apply more complex analyses that account for multiple levels of perceptual and linguistic abstraction; Section \ref{sec:integ} looks at work that applies large-scale ``integrative benchmarking'' to establish patterns of performance across many language models and neural response datasets; Section \ref{sec:comp_control} presents recent work on using language models to apply controls at the computational level rather than at the level of stimuli; Section \ref{sec:eval_improve} shifts to work that aims to evaluate and improve language models using insights and data from neurolinguistics, and finally, Section \ref{sec:conc} presents a discussion and outlook.

\section{Preliminaries}
\label{sec:prel}
\begin{figure}[t!]
\centering
\includegraphics[scale=0.45]{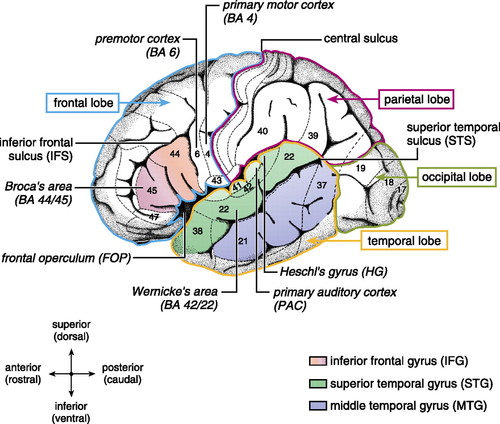}
\caption{Major language relevant gyri and Brodmann areas in the Left Hemisphere. Figure from \cite{friederici2011brain}.}
\label{fig:map}
\end{figure}

\paragraph{Language Models}
We take ``language model'' to mean any computational model that aims to explain and make predictions about some aspect of language. This includes early models of syntactic structure \citep{chomsky1956three, pollard1988information, bresnan2015lexical}, lexical distributional models \citep{schutze1993word, mikolov2013efficient, pennington2014glove}, and more recent ANN models \citep{hochreiter1997long, vaswani2017attention}. 
\paragraph{Neuroimaging methods}
Work surveyed in this paper makes use of the following neuroimaging modalities:

\begin{itemize}
    \item \textbf{Electroencephalogram (EEG)} involves recording the electrical activity occurring in the cortex over a period of time using multiple electrodes placed on the scalp \citep{henry2006electroencephalography}. EEG is generally considered to have a high temporal but a relatively poor spatial resolution (centimeters) and is often used for deriving event-related potentials (ERPs) which average over the EEG signals resulting from of a specific event, e.g. reading a word \citep{luck2014introduction}.
    \item \textbf{Magnetoencephalography (MEG)} involves measurement of the magnetic field generated by the electrical activity of neurons in the cortex. Like EEG, MEG offers an accurate resolution of the timing of neuronal activity. Unlike EEG, it also offers a relatively good spatial resolution (millimeters) \citep{baars2013fundamentals}.  
    \item \textbf{Functional Magnetic Resonance Imaging (fMRI)} measures neuronal activity in the brain via blood oxygenation level-dependent (BOLD) contrast. fMRI offers high spatial resolution and signal reliability, but poor temporal resolution ($3$ to $6$ seconds) due to slow nature of the hemodynamic response \citep{soares2016hitchhiker}. 
    \item \textbf{Electrocorticography (ECoG)} records electrical activity in the brain through electrodes placed in direct contact with the surface of the brain \citep{baars2013fundamentals}. ECoG data has a fine spatial and temporal resolution and a high signal-to-noise ratio. As the procedure is invasive, however, ECoG data can only be gathered as part of a clinical procedure. 
\end{itemize}
\paragraph{The language network}  A  range of brain regions --- often refered to as regions of interest (ROIs) --- in the left and right hemisphere have been implicated in facilitating language. Although there is no consensus on the exact functional neuroanatomy, the network is known to, broadly speaking, include parts of the inferior frontal, the superior temporal, and the middle temporal gyra in the frontal and temporal lobes as well as the inferior parietal and angular gyrus in the parietal lobe \citep{friederici2011brain} --- see Figure \ref{fig:map}. For more details see  \cite{friederici2013language,poeppel2014neuroanatomic,fedorenko2014reworking, blank2016syntactic, pylkkanen2019neural, pylkkanen2020neural, fedorenko2020lack}. 

\paragraph{Linking hypotheses}
To relate the computational and neural recording paradigms, a linking hypothesis is assumed ---  a commonly employed hypothesis, for instance, is that brain activation magnitude should correspond to measures of processing difficulty or complexity derived from the stimuli. Another common linking hypothesis is that of linear mapping, where a linear model is trained to map between e.g. features extracted from a computational model and neural recordings of a set of stimuli, under the assumption that a linear transformation should suffice to model the relationship between the two spaces. Goodness of fit or predictive accuracy are used to evaluate correspondence between the two.

\section{Background}
\label{sec:old_studies}
Investigations into the neural basis of language processing aim to characterize the processes which occur once an utterance is perceived that enable a listener to arrive at a contextualised meaning from the sensory input she perceives. Such investigations date back at least to \cite{broca1861perte} and \cite{wernicke1874aphasische}'s description of two brain regions linked to the language faculty\footnote{Although the term is still commonly employed, it is now understood that Broca's area 'is not a natural kind' and instead consists of consists of multiple functionally distinct components \citep{fedorenko2020broca} only part of which are directly linked to language.}. In the 1970s, the introduction of non-invasive brain-monitoring techniques allowed scientists to characterize activity in the brain using methods such as EEG  and MEG. Using these methods, researchers initially identified Event-related potential components such as N400, P600, and ELAN\footnote{N400, related to semantic processing, is a negative-going potential, which peaks around 400ms after stimulus onset; P600, associated with syntactic processing, is positive-going potential peaking around 600ms after stimulus onset. ELAN is an early left anterior negativity, characterized by a negative-going wave that peaks around 200 ms or less after onset.} which corresponded to levels of syntactic or semantic processing \citep{kutas1980reading, kutas1984brain, hagoort1993syntactic, friederici1993event}.

While these early studies were successful at identifying particular patterns of activation which occurred during linguistic processing, they did not offer computational models which could make directly testable predictions of activation patterns. \cite{mitchell2008predicting} did this, demonstrating that word representations based on co-occurrence statistics could in fact be used to predict the activations associated with concrete nouns as measured via fMRI recordings. To accomplish this, they encode each stimulus word as a vector of semantic features computed from the its occurrences in a large text corpus. An linear encoding model is then trained to predicts fMRI
activation per voxel in the brain, as a weighted sum of the semantic features. Evaluation was carried out through “leave-two-out” cross-validation, in which a model is repeatedly trained
while holding out two word stimuli from the full set, then
tested on whether its predicted fMRI image for these two stimuli can 
select the correct one via cosine similarity.

\section{The Meanings of Words and Phrases}
\label{sec:meaning_nouns}

 Following a setup similar to \cite{mitchell2008predicting}:  \cite{murphy2009eeg} applied the same methodology using EEG instead of fMRI; \cite{devereux2010using} used four automatic feature extraction methods leveraging different sources of information from corpora; \cite{pereira2013using} employed low-dimensional representations constructed by applying topic modelling to a small wikipedia corpus, showing that this feature space can outperform the one used by \cite{mitchell2008predicting} in a classification task based on decoding the values of semantic features present in concepts from the fMRI data; \cite{palatucci2009zero} also reversed the original task, decoding word representations based on both co-occurance statistics and human annotations from their corresponding fMRI recordings in a zero-shot setting; \cite{sudre2012tracking} studied the temporal sequence of language processing, showing that perceptual and semantic features could be decoded at different times from MEG data; \cite{anderson2013words} used image-based distributional semantic representations of concepts instead of text-based ones; \cite{anderson2017visually} built on this, showing that although both perform equally well for concrete concepts, text-based word representations better predict the brain activations of abstract concepts compared to visually-grounded ones; \cite{bulat2017speaking} employed multiple evaluation methods to carry out a systematic appraisal of how well a wide range of text-based and grounded semantic models --- including more recent skip-gram and bag-of-words word embeddings \citep{mikolov2013efficient} --- can predict fMRI measuremets;  \cite{pereira2018toward} presented a new fMRI dataset of subjects reading words and passages, showing that a decoder trained to predict word embeddings from imaging data for individual concepts that were selected using a novel sampling procedure designed to cover the entire semantic space, can generalize to new concepts and can, further, decode sentence stimuli represented as a simple average of (content) word embeddings; finally, \cite{abnar2017experiential} evaluate eight types of word embeddings on how well they predict
the fMRI activation recordings from \cite{mitchell2008predicting}, finding that a word embedding method which incorporates syntactic information fares best compared to multiple other word representations base on the skip-gram approach, matrix factorization, or crowd-sourced association features.

In a notable critique, \cite{bullinaria2013limiting} present evidence that it is the lack of representational distinctiveness of the fMRI voxel activation vectors that is the major limiting factor in the kind of work described above. \cite{gauthier2018does} take aim at the evaluation methods employed in decoding studies where 'semantic representations' derived from stimuli are decoded from brain activations. They show that the evaluation techniques used in
these studies are underspecified and are therefore not able to distinguish between sentence representations drawn from models optimized for very different tasks. For using these models to make meaningful conclusions about the way linguistic processing
is realized in brain activity, they recommend: a) clear specification of the task and mechanisms in the brain hypothesized to be generating or consuming a given representation, b) breaking down the feature space into interpretable sub-spaces (see Section \ref{sec:comp_control} for examples of work where this is applied), and c) using encoding models to ablate the extent to which different model components can explain variance in neural response. Finally, \cite{beinborn2019robust} presented a standardized framework for brain-encoding studies, demonstrating the effect which choice of evaluation measure has on the interpretation of model predictive power. Based on this, they offered a set of recommendations regarding choice of metrics and reporting of results.

\section{Searching for Syntax}
\label{sec:syntax}
Concurrently, researchers have also sought to understand how the brain computes and represents syntactic structure during language comprehension. By varying the type of stimuli presented to subjects and carrying out comparisons between e.g. a) sentences of different complexities or b) sentences and lists of words, such studies were able to make conclusions about the brain regions which show sensitivity to sentence structure and about the temporal profile of brain activity \citep{just1996brain, dapretto1999form, humphries2006syntactic, hagoort2005broca, pallier2011cortical, fedorenko2012lexical, brennan2012time}.

The use of computational models in such work can be traced back to pyscholinguistic studies of syntactic processing difficulty where probablistic language models (such as probabilistic context-free grammars (PCFG)) were used to provide predictions about human reading times or grammaticality judgements \citep{christiansen1999toward, tabor1999dynamical, hale2001probabilistic, levy2008expectation, reitter2011computational}. \cite{parviz2011using} found that measures derived from an incremental syntactic parser and a 4-gram markov chain language model were predictive of the N400 ERP component. 

\cite{frank2013word} expanded on these findings, showing that word surprisal estimates from a Recurrent Neural Network language (RNN) model provided better predictions compared to n-gram models and phrase structure grammars. \cite{frank2015erp} went a step further, extracting four different word information (word and part-of-speech surprisal and entropy reduction) measures and evaluating their predictivity of six different ERP components which are known to be sensitive to violations. Their results indicated that readers’ expectations about upcoming words do not necessarily rely on hierarchical sentence structure (see \cite{frank2018hierarchical} for further relevant discussion). 

\cite{hale2015modeling} apply the same approach to neural
time courses obtained using fMRI. In contrast with \cite{frank2015erp}, their findings showed that grammatical predictors were predictive of BOLD (see Section \ref{sec:prel}) over n-gram baselines, indicating the sensitivity of human sentence processing to hierarchical structure, at least in the
anterior temporal lobe. They posited that this discrepancy might be due the
difficulty of measuring nuanced syntactic processing
activity with behavioral and ERP measures. \cite{brennan2016abstract} collected fMRI recordings of subjects listening to naturalistic storytelling data. Using this, they showed findings similar to those of  \cite{hale2015modeling}: abstract syntactic structures (context-free phrase structure grammars and and mildly context-sensitive X-bar structural descriptions) were predictive of brain activity in the temporal lobes, but not in other areas, whereas predictors derived from n-gram models showed correlations across a broad network of areas. 
\cite{brennan2017meg} built on this, using MEG data, they found that anterior temporal lobe (ATL) activity is well-predicted by a parse-step measure derived from a predictive left-corner parser, which is consistent with the hypothesis that the ATL is sensitive to basic combinatoric operations. \cite{henderson2016language} also arrived at similar findings. Using fixation-related fMRI\footnote{A method that combines eyetracking with BOLD to locate brain activity as a function of the currently fixated item.} and syntactic surprisal statistic derived from a PFCG, they found that this surprisal measure modulates activity in the left ATL and the left inferior frontal gyrus.  \cite{brennan2019hierarchical} further addressed the extent to which hierarchical structure is needed for language comprehension, using a naturalistic EEG dataset of participants listening to a chapter of Alice in Wonderland. They corroborate that, for left-anterior and right-anterior electrodes after around 200 ms from onset, syntactic surprisal measures derived from models which condition on hierarchical structure capture variance beyond models that condition on word sequences alone. 

Zooming in on the algorithmic level, \cite{stanojevic2021modeling} employed Combined Categorical Grammar (CCG) \citep{steedman2011combinatory}, a mildly context-sensitive grammar, to extract node count based complexity metrics. They found that these metrics could better predict fMRI activation time courses in language ROIs compared to Penn
Treebank-style context-free phrase structure grammars, confirming \cite{brennan2016abstract}'s finding that mildly context-sensitive grammars can better capture aspects of human sentence processing. They attribute this to an operation designed to make CCG handle “movement” constructions in a more plausible manner compared to CFGs. 

\cite{hale2018finding} pioneered the use of Recurrent neural network grammars
\citep{dyer2016recurrent} (RNNGs) as models of human syntactic processing. RNNGs are generative models of both
trees and strings (jointly) where neural networks are parametrized to choose parsing transitions. Using the same linking hypotheses described above and a beam search procedure (over derivations), they found that the measures estimated from the RNNG's intermediate states predicted several ERP components including P600. These same components were not significantly predicted using an LSTM \citep{hochreiter1997long} language language model which operates sequentially, having no access to structural information. \cite{brennan2020localizing} build on this, using RNNGs in conjunction with fMRI data recorded for the same Alice in Wonderland chapter to localise the information used for predictive processing across six ROIs associated with language processing. Their findings, confirm earlier ones: word-by-word surprisal\footnote{Note that surprisal here and in \cite{hale2018finding} is not the same as the syntactic surprisal measure based on part-of-speach category which is used in most previous studies, but is based on the lexical item itself.} derived from a sequential LSTM language model correlates with activity in a range of temporal and frontal brain regions; when surprisal is conditioned by explicit hierarchy as in the RNNG language model, fit is improved above the LSTM model in the left posterior temporal lobe and the left inferior parietal lobule. In addition, a measure based on the number of parsing steps explored by the RNNG model between words --- which is hypothesized to reflect compositional processing --- is found to be predictive of activation in most ROIs, particularly when the RNNG is setup to directly compose phrases, and not only encode hierarchy.

Most recently, \cite{reddy2021syntactic} proposed a shift from effort-based metrics (node count, surprisal, etc.) that allow for the localisation in the brain of a general notion of syntax to a methodology which allows for the studying of specific syntactic features. To accomplish this, they presented a subgraph embeddings-based method that models the constituency tree based structure of sentences. They showed that this method was more predictive of brain activity than a commonly used effort-based metric (node count), and used it to demonstrate that the brain encodes complex, phrase-level syntactic information. 

\section{Modelling Multiple Levels of Abstraction}
\label{sec:multiple}
Language comprehension from speech or text involves many perceptual and cognitive subprocesses, from perceiving individual words, to parsing sentences, to building semantic representations that are contextualised by world-knowledge and previous utterances in a discourse, etc. In this section, we survey work which has performed simultaneous analyses at multiple levels of perceptual and linguistic abstraction. 

In one of the first studies to use fMRI recordings from subjects reading naturalistic data (a chapter from \textit{Harry Potter and the Sorcerer's stone}), \cite{wehbe2014simultaneously} presented one such integrated analysis. Employing a  model consisting of a diverse set of visual, orthographic, lexical, syntactic, semantic, and discourse properties, they predicted neural activity per voxel as a linear combination of the features. Using this encoding (linear regression) model, they were able to examine the brain areas which were sensitive to the different types of features, enabling them to distinguish between areas on the basis of the type of information they represent. Using MEG data gathered for the same chapter of Harry Potter, \cite{wehbe2014aligning} was one of the earliest works to investigate the alignment between the representations used by RNN language models and brain activity as subjects read a story. They train auto-regressive neural language models (NLMs) \citep{mikolov2011rnnlm} on a corpus of Harry potter fan fiction and extract three classes of features per time-step: the embeddings, the hidden state vectors (previous and current), and the predicted output probabilities. In a series of classification experiments involving prediction of the MEG signal from each of these feature classes, they analyse how well each predicts MEG activity along the temporal and spatial dimensions, finding: a) brain activity is well predicted by the hidden state representation of (past) context, b) the embedding features are good predictors of the current word, and c) the activity in different brain regions is predicted with a delay that corresponds with the processing pathway that starts in the visual cortex
and moves up.  \cite{lopopolo2017using} followed a similar approach with the goal of disentangling phonological, lexical, and syntactic information present in the fMRI recordings of subjects listening to a set of naturalistic (Dutch) literary texts. Instead of a neural network language model, however, they estimate perplexity statistics from  trigram markov models (based on lexical forms, part-of-speech tags, and transcribed phonemes). Their results evidence a set of cortical networks that are separately activated for each of the three types of information, with no significant overlap between them. 

\cite{huth2016natural} also made use of a naturalistic dataset of spoken stories. They represented each word in the stories as a 985-dimensional vector built from co-occurrence statistics, intended to encode semantic information. A linear regression model was estimated per voxel, from the word representations. Controlling for lower-level features, they evaluated correlation between predicted and actual BOLD signal, finding significant predictiveness in various areas associated with the brain's semantic network. They propose a Bayesian algorithm that constructs a generative model of areas tiling the cortex across subjects, resulting in single atlas that describes the distribution of semantically selective functional areas in human cerebral cortex. \cite{jain2018incorporating} follow \cite{wehbe2014aligning} in using an RNN language model to incorporate context into encoding models that predict the neural response (fMRI in this case) of subjects listening to natural speech. They find that the representations from NLM hidden states outperform previously used non-contextual word embedding models in predicting brain response and that context length and choice of layer differentially predict the activation across cortical regions.  \cite{lebel2021voxelwise} followed a similar approach, focusing on the Cerebellum. Using features that span the hierarchy of language processing, they showed that neural response to language in the Cerebellum is
best explained by high-level conceptual features --- especially those associated with social semantic categories --- rather than lower-level acoustic or phonemic ones. Finally, \cite{li2020modeling} formalised several linguistic theories about pronoun resolution as symbolic computational models, evaluating them as well as an ANN conference resolution model according to their ability to predict brain activity patterns (both MEG and fMRI) time-locked at each third person pronoun in 'The Little Prince' dataset as English and Chinese subjects listened to an audiobook recording \citep{stehwien2020little}. They find that the memory-based ACT-R model \citep{van2013wm}, which chooses the entity in working memory with the highest activation as the antecedent of a pronoun, best explains the neural response associated with pronoun resolution.

 In seminal work adjacent to that described above because it studies neural oscillation patterns, \cite{martin2017mechanism} showed that a computational model \citep{doumas2008theory}, which uses time to encode hierarchy and was originally designed for relational reasoning, can be applied to sentence processing, exhibiting oscillatory patterns of activation closely resembling the human cortical response to the same stimuli. From this model which learns and generates structured, symbolic representations using time-based binding in a layered neural network, they are able to derive an explicit computational mechanism\footnote{See discussion of \cite{gauthier2018does} in Sec. \ref{sec:meaning_nouns}.}) for how the human brain might convert perceptual features into hierarchical representations, offering a linking hypothesis between linguistic and cortical computations. \cite{martin2020compositional} built on this, proposing an integrated model of language comprehension across multiple timescales starting from perception and lexicalisation to syntactic composition and comprehension. Drawing on ideas from models of entertainment to speech, structure building mechanisms from systems neuroscience (coordinate transforms via gain modulation), and the neurosymbolic relation learning model referenced above, the proposed model is able to unify the neurphysiological, cognitive, and linguistic computational levels.

\section{Integrative Benchmarking and Computational Convergence} \label{sec:integ}
While the shift from using several separate models which operate at different levels of abstraction to synergistic neural models has led to both better prediction of neural response and better performance on a range of linguistics tasks, \cite{schrimpf2020neural} posited that in order to understand the relationship between these computational models, neural response, and behaviour, large-scale integrative benchmarking is needed wherein patterns of performance are established across many models and datasets. Using measurements from three datasets\footnote{With stimuli that varied in length and domain and was either presented to subjects as audio or read.} of neural response recordings (fMRI and ECoG) as well self-paced reading data, they tested a wide range of computational models, from simple word embeddings to larger, recurrent and self-attention based ones, evaluating them based on how well they predicted the neural response recordings and the self-paced reading patterns, with reference to how well they perform tasks such as next word prediction. Their results demonstrated that: a) there is a variance across models in ability to predict neural response and self-paced reading patterns (e.g. GPT-2 \citep{radford2019language} almost completely explains the variance in neural response, while GloVe performs poorly), b) there is a consistency in how models score across datasets and experiments, c) models that perform better at next word prediction (but not the GLUE suite of natural language understanding tasks \citep{wang2018glue}) better predict neural response measurements and self-paced reading times, d) models that better predict neural response better predict reading times, and e) for some models , architecture alone, randomly initialised, can reliably predict brain activity and reading times. 

Also aiming to establish a systematic ontology of whether and when ANN models representations align with brain activations, \cite{caucheteux2020language} trained $7,400$ different ANN models with different architectures and objectives. They evaluated the models according to how well a ridge-regularised linear could be trained to map from their internal representations to MEG recordings of 104 subjects reading words sequentially presented randomly word or as sentences. Aiming to arrive at a spatio-temporal decomposition of the reading network, they found: a) as previously shown, the final-layer activations a deep convolutional neural network trained on character recognition are predictive of activation in the early visual cortex, b) word-type embeddings (Word2Vec \citep{mikolov2013efficient}) predicted brain response above and beyond the visual representations starting from $\approx$ 200 ms after onset, in the left-lateralized temporal and prefrontal cortices, especially, c) After $\approx$ 1 second from word onset, contextualised word representations from LMs led to significantly better prediction than the previous two feature sets, particularly in the regions associated with high-level sentence processing. Subsequently, \cite{caucheteux2021gpt} found that GPT-2's predictiveness of fMRI activation for a subject-story pair in the Narratives dataset correlated to subjects’ comprehension scores assessed per story.

Most recently, \cite{antonello2021low} adapted a encoder-decoder method from computer vision that measures transferability between different models to construct a language representation embedding space. Using this, they could describe and visualise the relationships between representations derived from a $100$ diverse types of language models, ranging from static word-embeddings (GloVe, etc.) and interpretable tagging models (part-of-speech, named entity recognition, chunking etc.) to machine translation models and autoregressive LMs (GPT-2, etc.) or masked LMs (BERT, etc.).  They show that this space has a low-dimensional structure and that it intuitively models how different representations relate to one another. Moreover, fitting encoding models to predict fMRI data from each of the $100$ language representations, they find the embedding space's structure, when mapped to the brain, reflects well-known language processing hierarchies and predicts which representations map
well to which areas in the brain. 

\section{Computational Controls}

\begin{figure*}[t!]
\centering
\includegraphics[scale=0.34]{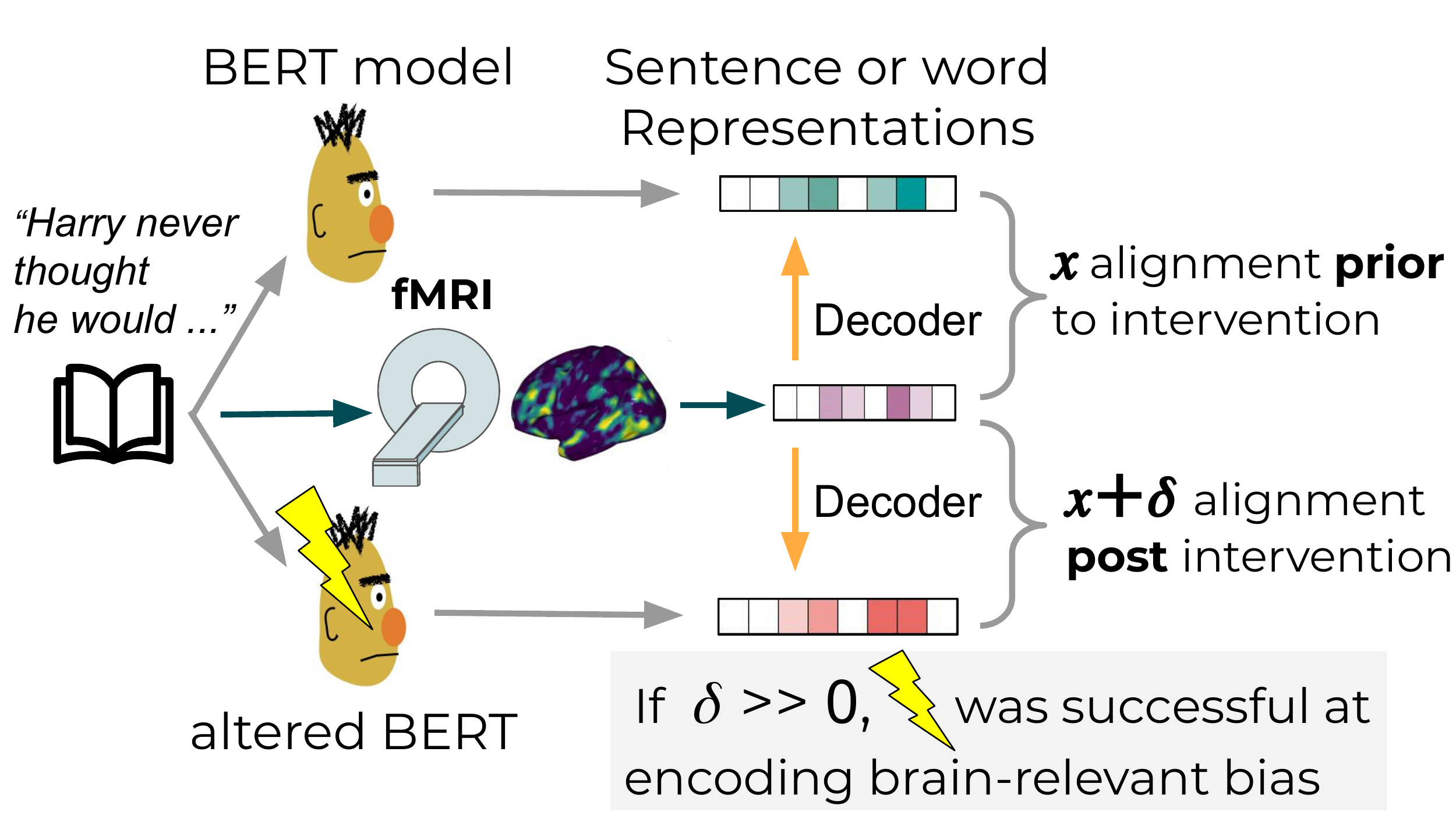}
\caption{An example of the application of computational control. A baseline BERT model and an 'altered' one are used to generate linguistic representation of stimuli. The intervention to alter the LM can be evaluated based on how well the resulting representations can be decoded from brain activation data compared to the baseline, controlling for other factors.}
\label{fig:control}
\end{figure*}

\label{sec:comp_control}
To establish a relationship between linguistic function and neurobiology, it is necessary to decompose the various facets of linguistic processing and map them onto the units of neurobiology \citep{poeppel2012maps}. Traditionally, this has been attempted through careful manipulation of stimuli e.g. scrambled sentences vs. natural sentences. Instead of applying strict controls at the level of the stimuli, recent work has explored the possibility of applying computational controls (see Figure \ref{fig:control} for a demonstrative example). When successful, this allows for a) testing a wider range of more specific questions concerning conditions which would be significantly more difficult or expensive to control for at the level of the stimuli and b) the use of naturalistic stimuli which more closely resemble real-world contexts.

For instance, by varying the length of context which is fed to a LM or providing it with distorted context,  \cite{jain2018incorporating} were able to ablate and the effect contextual information on alignment to neural response. \cite{abnar-etal-2019-blackbox} followed a similar protocol, employing an extension of the commonly used Representation Similarity Analysis paradigm \citep{kriegeskorte2008representational} where different instances of the same model are compared as a single parameter is altered. Varying the parameter of context length for four classes of language models, they find that increased context length not lead to increased representational alignment to brain recordings. 

\cite{gauthier2019linking} took a pretrained masked language model (BERT \citep{devlin2018bert}) as a baseline sentence representation model, finetuning it on a suite of tasks which are commonly used to evaluate 'Natural Language Understanding' (NLU). Training a regularised linear decoder to map from the fMRI sentence-level data of \cite{pereira2018toward} to representations extracted using the finetuned models, they find decreased brain decoding performance across all tested NLU tasks, but find improved decoding performance for scrambled language modelling tasks where fine-grained syntactic information is removed. This result might be seen as disagreeing with an emerging consensus in the literature regarding the role of hierarchical structure in sentence processing \footnote{e.g. \cite{hale2015modeling,brennan2016abstract, henderson2016language}, etc. but perhaps agreeing with \cite{frank2013word} and \cite{frank2015erp}. Refer to Section \ref{sec:syntax}.} and once again raised questions regarding the fine-grained-ness of the information which can be expected to be found in fMRI data using linear mapping as a linking hypothesis. \cite{abdou2021does} followed a similar methodology, proposing an approach which enables the evaluation of more targeted hypotheses about linguistic composition and structure by finetuning LMs with an auxilliary attention constraint to inject structural bias derived from three linguistic formalisms into LM representations.  They showed that, across the three formalisms, this improves brain decoding performance for the Harry Potter data, while for the \cite{pereira2018toward} data the effect is less clear.

The methods outlined so far apply controls by manipulation of either the input or the training objective and task. \cite{toneva2020combining} devised an approach for studying the neural substrates associated with the 'supra-word meaning' of a phrase with a computational control that disentangles composed-from individual-word meaning. Using regularised linear regression models, they learn mappings between word embeddings and context representations, which they then subtract from the representations of items in the target space to obtain 'residual' representations for a word (sans contextual information) or for a phrase \textit{beyond} its individual words. For an fMRI dataset, they find that this supra-word representation predicts  activity in the anterior and posterior temporal cortices. In MEG recordings, however, they find no clear signal for supra-word meaning. Similarly, \cite{caucheteux2021disentangling} proposed a method to factorise distributed LM representations according to a taxonomy of:  \textit{syntax vs. semantics} and \textit{lexical vs. compositional}. They constructed syntactic representations by averaging over the LM activations for a set of sentences with the same syntactic structure. A LM's word embedding is taken as a lexical representation, and the contextualised representation of higher layers as the compositional one. A semantic representation is taken to be the residual
of subtracting the syntactic representation, which can be done at the lexical or the compositional level. Extracting each of these representations from the relevant stimuli text, they learn ridge-regularised mappings to fMRI recordings of subjects listening to the stories from the Narratives dataset \citep{nastase2021narratives}. Their results showed: a)  compositional representations
recruit a broader cortical network than
lexical ones and b) syntax and semantics
appear to share a common 
neural basis, in line with recent findings from the neuroscience literature \citep{fedorenko2020lack}. 

Examining the question of predictive coding \citep{rao1999predictive} in language processing, \cite{goldstein2021thinking} investigated whether (and how) humans and LMs engage in spontaneous prediction when processing language. To accomplish this, they: a) used a sliding-window method to show humans could effectively predict the next word in a transcribed story across sentence positions and part-of-speech; b) extracted word-by-word prediction probabilities from GPT-2 for the same story, finding them to be correlated to the human predictability scores, with better correlations as the amount of previous context fed to  the model increased; c) showed, in a set of experiments with ECoG data recorded while subjects listened to a spoken story, that static word embeddings (GloVe) as well as an arbitrary embeddings baseline designed to ablate the effect of correlations between adjacent word embeddings of bigrams, could predict neural activation (across electrodes in the left hemisphere) starting from 800 ms before word onset; d) at earlier time from word onset, neural responses were better predicted for predictable words than for unpredictable words, e) neural response  before onset was better modelled for the words subjects predicted even when they did not match the correct words which they subsequently perceived; f) employing representations which incorporate previous context from GPT-2's activations led to both better and earlier (i.e. pre-onset) prediction of neural activity, particularly in high-order language areas; g) ablated this result, showing that it was due to both better representation of previous context (compared e.g. to a baseline of mean-pooled GloVe embeddings) and to improved next-word prediction.
Finally, \cite{jain2021interpretable} presented a novel multi-timescale encoding model for predicting fMRI responses to natural speech. Using a) an LSTM where the memory timescale of each individual unit is fixed according to a power law distribution and b) a Gaussian radial basis function kernel to downsample the stimuli representations, they are able directly estimate the timescale of information represented in a voxel of fMRI data from the encoding model's weights. They find that this method which relies on 'computational control' leads to a more fine-grained map of timescale selectivity compared to previous work which relied on stimulus manipulation \citep{lerner2011topographic}. 

\section{Using Brain Activation Measurements to Improve/Evaluate NLP Models}
\label{sec:eval_improve}
The work described in this survey so far has employed computational language models towards the study of human language processing. Working in the other direction, researchers have also attempted to leverage data and insights from neurolinguistics to evaluate and improve language models. To that end, \cite{fyshe2014interpretable} presented an algorithm that integrates brain activations into the construction of word-type vector space models. Their approach, based on Non-Negative Sparse Embeddings \citep{murphy2012learning}, constrains the embedding space so that words close in brain activation space also have similar representations in the embedding space. They find the resulting embeddings to better match a behavioral measure of semantics and to better predict corpus data for unseen words. In an opinion paper on word embedding evaluation, \cite{sogaard2016evaluating} presented some preliminary experiments and argued that because evaluation on downstream tasks is expensive and impractical and evaluation on corpus statistics is circular, alignment to human word processing measurements should be used as a an evaluation approach. On that account, \cite{hollenstein-etal-2020-cognival} presented CogniVal, a framework for the evaluation of word embeddings based on their ability to predict data from 15 datasets of eyetracking, EEG, and fMRI signals recorded during language processing.

Testing the possibility of using EEG data to improve NLP models, \cite{hollenstein2019advancing} extract word-level EEG features from different frequency ranges, using the ZUCO dataset \citep{hollenstein-etal-2020-zuco} which contains simultaneous eye-tracking and EEG measurements of natural sentence reading. Combining these features with standard word-level and character-level information employed by NLP models, they find (modest) improvements over baselines which do not include the EEG features across three tasks (named entity recognition, relation classification and sentiment analysis). 

Pioneering the use of LMs for \textit{in silico} modelling\footnote{Where a computational model is used to simulate (some facet of) brain function.}, \cite{schwartz2019inducing} finetuned BERT models to predict (fMRI and MEG) brain activity measurements, biasing them to learn generalizable relationships between text and brain activity. They find that the the fine-tuned models are better at predicting brain activity across subjects and recording modalities than non-finetuned models. Finally, \cite{toneva2019interpreting} used brain response data to interpret the information encoded in the internal representations of different layers of LMs for various context lengths, based on how well they predicted activation in different groupings of language ROIs.  Ablating the role of attention, they found that replacing the learned attention function with a uniform distribution in early BERT layers led to better prediction of brain activation. Applying these altered models to a task which evaluates the syntactic capabilities of LMs \citep{marvin2018targeted}, they found significant improvements compared to the vanilla pretrained model, demonstrating that altering models so that they align better 
with brain recordings can lead to better performance on NLP benchmarks. 

\section{Outlook}\label{sec:conc}
Present-day advances in machine learning have enabled the building of language models which through training on immense volumes of data are capable of simulating human behaviour on various linguistic tasks better than ever before. Concurrently, datasets of neural response recordings that both include more subjects and utilise a larger amount of naturalistic stimuli than previously possible have been made openly accessible to the research community \citep{bhattasali-etal-2020-alice,stehwien2020little, nastase2021narratives}. Exploiting these advances, recent studies a) found evidence that autoregressive ANN language models converge to solutions that reliably align to brain activations \citep{caucheteux2020language, schrimpf2020integrative} and b) worked towards furthering our understanding of fundamental aspects of naturalistic language comprehension, e.g. the role of hierarchical structure \citep{brennan2019hierarchical, stanojevic2021modeling}, the function of predictive coding \citep{goldstein2021thinking}, and the neural substrates responsible for lexical and combinatorial semantics \citep{toneva2020combining}. 

While the manner in which current ANN language models learn is manifestly inefficient and un-human-like --- leveraging text-only information and requiring orders of magnitude more data than a human child --- the solutions to which they converge have been shown to both remarkably simulate aspects of linguistic understanding and align to cognitive measurements. This offers reasonable grounds for optimism that the time is ripe for these models to both contribute to and benefit from work on the \textit{Mapping Problem} \citep{poeppel2012maps}: the problem of mapping the elementary units of linguistic processing to their neurobiological counterparts. As \cite{hale2021neuro} argues, linguistically-interpretable models are likely to be key for this symbiosis, allowing for a principled decomposition of a model's components into smaller linguistically meaningful units. In view of the approaches to computational control described in Section \ref{sec:comp_control}, I posit that even models that are not intrinsically interpretable will be useful, given the plethora of interpretability methods recently developed \citep{belinkov-etal-2020-interpretability}.  

By serving as a plausible simulation of human learning, future work that explores more human-like training setups (e.g. multi-modal or embodied learning), objectives, and model architectures can also help empirically answer long-standing questions in neuro- and cognitive linguistics.


\bibliography{anthology, custom}
\bibliographystyle{acl_natbib}


\end{document}